\documentclass[journal]{IEEEtran}
\usepackage{cite}
\usepackage[pdftex]{graphicx}
\usepackage[cmex10]{amsmath}
\usepackage{amssymb}
\usepackage{algorithm,algorithmic}
\usepackage{url,color}

\usepackage[T1]{fontenc} 
\usepackage{balance}
\usepackage{multirow}
\usepackage{subcaption}

\usepackage{array}
\usepackage{colortbl}

\newcommand{\ie}{\emph{i.e.}}
\newcommand{\eg}{\emph{e.g.}}

\newcommand{\diff}[2]{\frac{\partial #1}{\partial #2}}
\newcommand{\vct}[1]{\ensuremath{\boldsymbol{#1}}} 
\newcommand{\mat}[1]{\ensuremath{\mathbf{#1}}}
\newcommand{\set}[1]{\ensuremath{\mathcal{#1}}}

\newcommand{\tsum}{\ensuremath{\textstyle \sum}}

\newcommand{\T}{\ensuremath{\top}}

\newcommand{\argmin}{\operatornamewithlimits{\arg\,\min}}

\newcommand{\myparagraph}[1]{\vspace{0.1pt} \noindent \textbf{#1}}

\newcommand{\ssfm}{SSVM}
\newcommand{\svm}{SVM}
\newcommand{\svmsel}{SVM-sel}
\newcommand{\svmred}{SVM-red}

\begin{document}
\title{Super-sparse Learning in Similarity Spaces}
\author{
	Ambra~Demontis~\IEEEmembership{Student Member,~IEEE,}
	Marco~Melis~\IEEEmembership{Student Member,~IEEE,}\\
	Battista~Biggio~\IEEEmembership{Member,~IEEE,}
        Giorgio~Fumera~\IEEEmembership{Member,~IEEE,}
        and~Fabio~Roli~\IEEEmembership{Fellow,~IEEE}
\\Department of Electrical and Electronic Engineering,\\
University of Cagliari, Piazza d'Armi, 09123 Cagliari, Italy\\
\thanks{Published on: IEEE Computational Intell. Mag., 11(4):36-45, Nov 2016.}
\thanks{http://ieeexplore.ieee.org/document/7587466/}
\thanks{}
\thanks{A. Demontis: e-mail ambra.demontis@diee.unica.it}%
\thanks{M. Melis: e-mail marco.melis@diee.unica.it}%
\thanks{B. Biggio (corresponding author): e-mail battista.biggio@diee.unica.it}%
\thanks{G. Fumera: e-mail fumera@diee.unica.it}%
\thanks{F. Roli: e-mail roli@diee.unica.it}
}

\maketitle

\begin{abstract}
In several applications, input samples are more naturally represented in terms of similarities between each other, rather than in terms of feature vectors. In these settings, machine-learning algorithms can become very computationally demanding, as they may require matching the test samples against a very large set of reference prototypes. To mitigate this issue, different approaches have been developed to reduce the number of required reference prototypes. Current reduction approaches select a small subset of representative prototypes in the space induced by the similarity measure, and then separately train the classification function on the reduced subset. However, decoupling these two steps may not allow reducing the number of prototypes effectively without compromising accuracy. We overcome this limitation by jointly learning the classification function along with an optimal set of virtual prototypes, whose number can be either fixed a priori or optimized according to application-specific criteria. Creating a super-sparse set of virtual prototypes provides much sparser solutions, drastically reducing complexity at test time, at the expense of a slightly increased complexity during training. A much smaller set of prototypes also results in easier-to-interpret decisions. We empirically show that our approach can reduce up to ten times the complexity of Support Vector Machines, LASSO and ridge regression at test time, without almost affecting their classification accuracy.
\end{abstract}

\section{Introduction}
\label{sect:introduction}

In a growing number of applications, including computer vision, biometrics, text categorization and information retrieval, samples are often represented more naturally in terms of similarities between each other, rather than in an explicit feature vector space~\cite{chen09-jmlr,pekalska01-jmlr}.
Traditional machine-learning algorithms can still be used to learn over similarity-based representations; \eg, linear classification algorithms like Support Vector Machines (SVMs)~\cite{vapnik95-book,vapnik95} can be trained in the space implicitly induced by the similarity measure (\ie, the kernel function) to learn nonlinear functions in input space.
However, the main drawback of similarity-based techniques is their high computational complexity at test time, since computing their classification function often requires matching the input sample against a large set of reference prototypes, and evaluating such similarity measures is usually computationally demanding.
Even SVMs, that induce sparsity in the number of required prototypes (the so-called support vectors, SVs) may not provide solutions that are sparse enough, as the number of prototypes (\ie, SVs) grows linearly with respect to the number of training samples~\cite{steinwart03,chapelle07}.
To reduce the number of reference prototypes, several state-of-the-art approaches select them from the training data, and then separately train the classification function using the reduced set of prototypes.
However, decoupling these two steps may not effectively reduce the number of prototypes, without significantly affecting classification accuracy~\cite{duin06,pekalska01-jmlr,scholkopf99}. 

In this work, we first discuss the relationship between current prototype-selection methods and 
our recently-proposed approach for learning super-sparse machines on similarity-based representations~\cite{biggio15-icb,demontis15-iciap,melis15-iciap}. We then show that our approach can successfully tackle this issue by jointly learning the classification function along with an optimal set of virtual prototypes. The number of prototypes required by our approach can be either fixed a priori, or optimized through a carefully-designed, incremental cross-validation (CV) procedure.
Creating a super-sparse set of virtual prototypes allows us to learn much sparser solutions, drastically reducing the computational complexity at test time, at the expense of a slightly increased computational complexity during training.
A much smaller set of reference prototypes also provides decisions that are easier to interpret.
We validate our approach on two application examples, including biometric face verification and age estimation from faces. Our approach does not almost affect the generalization capability of SVMs, LASSO~\cite{tibshirani96-lasso}, and ridge regression~\cite{hoerl70-ridge}, while being capable of reducing their complexity of more than ten times, overcoming the performance of other reduction methods.
We conclude this paper with a discussion of future research directions.

\section{Learning in Similarity Spaces} 
\label{sect:learning-similarity-spaces}

Two main approaches can be used to learn classification and regression functions in similarity spaces, \ie, functions that depend only on similarities between samples, and not on their features~\cite{pekalska01-jmlr,chen09-jmlr}.
The first one consists of computing an explicit representation of samples in a vector space that aims to preserve the similarity values between the original samples. 
A well-known example is related to kernel functions (\ie, positive semi-definite similarity functions), as one may exploit the eigenvalue or Cholesky decomposition of the kernel matrix to represent samples in an explicit vector space, called the empirical kernel mapping~\cite{scholkopf99,pekalska01-jmlr,duin06,chen09-jmlr,pekalska09-tpami}.
Their underlying idea is to decompose the kernel (similarity) matrix $\mat K \in \mathbb R^{n \times n}$ computed on $n$ training samples as $\mat K = \mat X^{\T} \mat X$, where $\mat X \in \mathbb R^{n \times n}$ represents the training samples in the empirical kernel space. Then, at test time, never-before-seen samples should be mapped onto the same space, to be classified in a consistent manner, and this often requires matching them against all training samples~\cite{pekalska01-jmlr,chen09-jmlr}.
Clearly, the aforementioned decompositions are only possible for positive semi-definite similarities (\ie, kernel matrices).
For indefinite similarities different techniques can be exploited to account for the negative eigenvalues of $\mat K$, and potentially exploit the same decompositions, or adapt the corresponding classification functions~\cite{pekalska01-jmlr,chen09-jmlr,pekalska09-tpami}.
They include: spectrum clip, in which the negative eigenvalues are set to zero;
spectrum flip, in which their absolute values are used; spectrum shift, in which they are increased by a quantity equal to the minimum eigenvalue, such that the smallest one becomes exactly zero;
and spectrum square, in which the eigenvalues are squared~\cite{chen09-jmlr}. 
Notably, spectrum flip amounts to mapping the input data onto a pseudo-Euclidean space. This space consists of two Euclidean spaces: one for which the inner product is positive definite, and one for which it is negative definite.
This enables computing quadratic forms (\eg, vector norms and inner products) as the difference of two positive-definite norms.
Pseudo-Euclidean spaces are a particular case of finite-dimensional Kre\u{\i}n spaces consisting of two real subspaces. In general, indefinite kernels and similarities allow a consistent (\eg, infinite-dimensional) representation in a Kre\u{\i}n space~\cite{pekalska01-jmlr,chen09-jmlr,pekalska09-tpami}.

\begin{figure*}[t]
\linespread{1}
 \centering
  \includegraphics[scale=0.9]{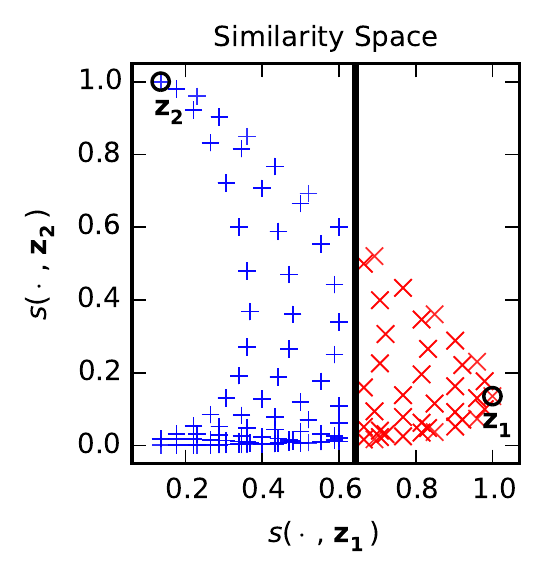}  
 \includegraphics[scale=0.9]{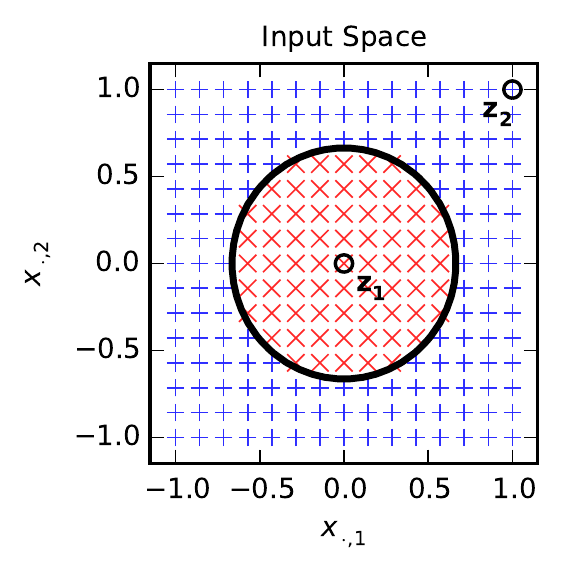}
 \caption{\emph{Left:} A linear SVM classifier trained in a two-dimensional similarity space, using similarities as  features, $\vct z_{1}$ and $\vct z_{2}$ as the reference prototypes, and the RBF kernel $s(\vct x, \vct z) = \exp \left( - \| \vct x - \vct z \|^{2}\right )$
as the similarity measure. Blue `$+$' and red `$\times$' represent the training data, while the black solid line denotes the classification boundary.
\emph{Right:} The corresponding nonlinear decision function in the input space.}
\label{fig:2D-sim_space-example}
\end{figure*}

The second approach consists of learning classifiers directly in the similarity space, \ie, exploiting similarities computed against a set of reference prototypes as feature values. 
This is equivalent to the former approach, if the spectrum-square technique is used~\cite{chen09-jmlr,pekalska01-jmlr} (see the example in Fig.~\ref{fig:2D-sim_space-example}). 
Worth remarking, some learning algorithms have been explicitly modified to deal with similarities, instead of adapting the similarity-based representation to existing learning algorithms. Examples can be found in the area of relational fuzzy clustering~\cite{hathaway94-pr}, and relational lexical variant generation~\cite{seaghdha09,pedersen07}.

Feature- and similarity-based representations may be thought as two facets of the same coin: modifying the similarity measure amounts to modifying the implicit feature space in which the linear decision function operates, and vice versa.
This means that, to achieve good generalization capabilities, it is necessary to properly define this space on the basis of the given learning algorithm.
When one is given a similarity-based representation, this space can be manipulated essentially in two ways, \ie, by either modifying the similarity measure or the prototypes.
Several approaches have been proposed to manipulate the similarity measure, including multiple kernel learning and similarity learning~\cite{sonnenburg06,chechik10}. They exploit a parametric similarity or distance measure, whose parameters are tuned by running a learning algorithm on the training data. In particular, in the case of multiple kernel learning, the goal is to learn the coefficients of a convex linear combination of a set of given kernels~\cite{sonnenburg06}.
Conversely, only few works have addressed the problem of selecting the reference prototypes to reduce complexity of similarity-based classifiers~\cite{duin06},
especially in the context of structured inputs like graphs and strings~\cite{riesen07,spillmann06-spr}.
To this end, it is also worth remarking that the Nystr\"om approximation can be exploited to approximate $n \times n$ similarity matrices based on a subset of $m \ll n$ randomly-chosen prototypes, reducing the complexity of computing all pairwise similarities from $\set O(n^{2})$ to $\set O(n m^{2})$. This approximation also works for indefinite kernels and similarities~\cite{gisbrecht10-nips}.

The latter approach, based on learning classifiers directly in the similarity space, includes the solution we propose in this paper and it is of particular interest in applications where the similarity metric is: ($i$) defined and can not be modified, ($ii$) not given in analytical terms, and ($iii$) not necessarily positive semi-definite.
For instance, in fingerprint recognition, the similarity measure is often defined a priori, as it encodes the knowledge of domain experts, and in most of the cases it is not positive semi-definite (it does not obey the triangle inequality).
In addition, it is usually computed by a physical device, called matcher, and it is not even analytically defined. 
Modifying the similarity measure in these cases is not possible.
The only way of manipulating the space induced by such a measure consists of modifying the prototypes.
The main limit of the corresponding state-of-the-art approaches 
is however intrinsic to the fact that they separately select the 
prototypes and then learn the classification function~\cite{duin06}.
The approach advocated in the next section, instead, is based on the idea of jointly optimizing the prototypes and the parameters of the classification function, to outperform existing reduction methods in similarity spaces.
Furthermore, with respect to methods devoted to reduce the number of SVs in SVMs, our approach requires neither the similarity to be a positive semi-definite kernel, nor the learning algorithm to be an SVM~\cite{biggio15-icb}. It can be applied, in principle, to reduce the complexity of any similarity-based learning algorithm.

\section{Super-Sparse Virtual Vector Machines} 
  \label{sect-super-sparse-virtual-vector-machine}

We present here our approach to learn super-sparse machines, inspired from~\cite{biggio15-icb,demontis15-iciap,melis15-iciap}.
The underlying idea is to reduce complexity of similarity-based learning algorithms by employing a very small  set of virtual prototypes. The virtual prototypes obtained by our learning algorithm are not necessarily part of the training data, but are specifically created with the goal of retaining a very high generalization capability.

Let us denote with $\set D = \{ \vct x_{i}, y_{i} \}^{n}_{i=1}$ the training data, where $\vct x_{i} \in \set X$ are the input samples and $y_{i} \in \set Y$ are their labels. We consider here a vectorial representation of the input data $\vct x$, \ie, we assume that $\set X$ is a vector space.
The output space $\set Y$ depends on whether we are considering a regression or classification problem. For regression, we consider $\set Y \subseteq \mathbb R$, whereas for two-class classification, we set $\set Y=\{-1,+1\}$.
The set of virtual prototypes is denoted with $\vct z = \{\vct z_{j}\}_{j=1}^{m}$,
where $ m \ll  n$ to obtain much sparser solutions.
The value of $m$ can be either fixed a priori, depending on application-specific constraints (\eg,  specific requirements on storage and time complexity, as in match-on-board biometric verification systems), or it can be optimized through a well-crafted CV procedure, as defined at the end of this section.
Note that the virtual prototypes $\vct z_{j}$ belong to the same input space $\set X$ as the training samples. We finally denote the similarity function with $s: \set X \times \set X \mapsto \mathbb R$. It can be any symmetric similarity function (not necessarily positive semi-definite).

Our goal is to learn a discriminant function $g(\vct x)$ consisting of a very sparse linear combination of similarities computed against the virtual prototypes: 
\begin{eqnarray}
g(\vct x) = \tsum_{j=1}^{m} \beta_{j} \phi_{j}(\vct x) + b = \tsum_{j=1}^{m} \beta_{j} s(\vct x, \vct z_{j}) + b 
\enspace, 
\end{eqnarray}
where $\vct \beta = (\beta_{1}, \ldots, \beta_{m}) \in \mathbb R^{m}$ is the vector of coefficients (one per virtual prototype), and $b$ is the bias.
Our approach is specifically designed to work in the similarity space, \ie, our basis functions $\phi_{j}(\vct x)$ are similarity functions $s(\vct x, \vct z_{j})$.
The reason is that we aim to optimize the virtual prototypes $\vct z_{j}$ without changing $s$, as in several applications such a function can not be changed; \eg, in biometric verification, one is often given a matching function which is neither customizable nor even known analytically.

As in \cite{biggio15-icb,scholkopf99}, we consider a regression problem in which the goal is to minimize the distance between the target variables $y$ and $g$ on the training points, with respect to the parameters of $g$, \ie, the coefficients $\vct \beta$, $b$ and the set of virtual prototypes $\vct z$. We do not constrain the prototypes $\vct z$ to be in $\set D$, but enable the creation of novel (virtual) prototypes. This allows our algorithm to achieve a better trade-off between accuracy and the number of required prototypes.
\begin{figure*}[t]
\linespread{1}
\centering
\includegraphics[width=0.775\textwidth]{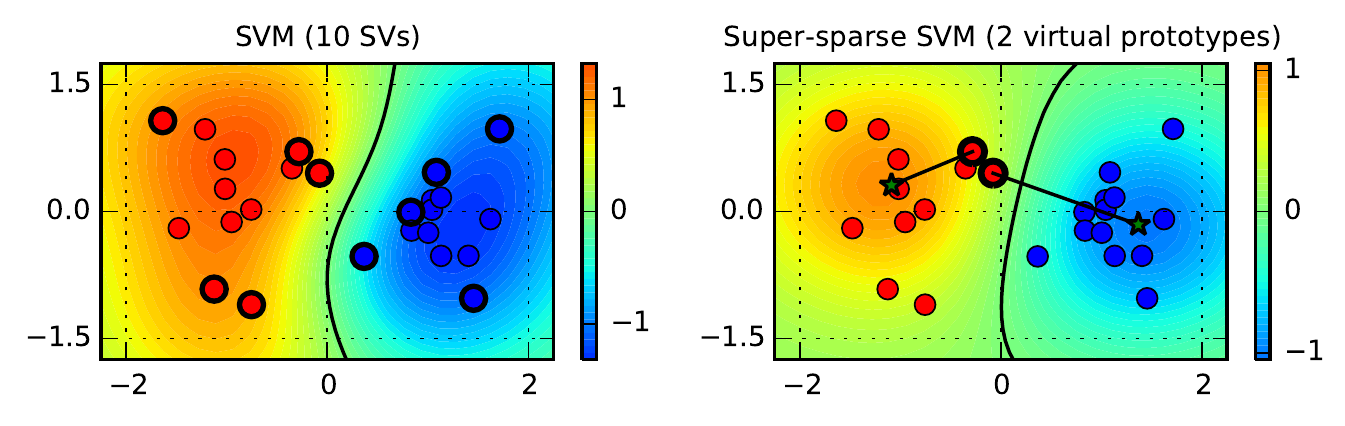}
\caption{A two-dimensional example showing the values of the discriminant function (in colors) of an SVM with the RBF kernel (\emph{left}) trained on 25 samples (red and blue points in the left plot), and of $g(\vct x)$ for our super-sparse learning algorithm (\emph{right}). Our approach almost exactly replicates the SVM's decision function (solid black line) using only 2 virtual prototypes instead of 10 SVs (highlighted with black circles in the left plot). The initial and final positions of the two virtual prototypes optimized by our approach are also shown in the right plot, respectively as black circles and green stars.}
\label{fig:2D-example}
\end{figure*}
The problem can be thus formulated as:
\begin{align}
\label{eq:objective}
\textstyle \min_{\vct \beta, b, \vct z} \;
\tsum_{i=1}^{n} u_{i} \left ( g(\vct x_{i})- y_{i} \right )^{2} + \lambda \vct \beta^{\T} \vct \beta
\enspace ,
\end{align}
where the scalars $u_{1},\ldots,u_{n}$ balance the contribution of each training sample $\vct x_{i}$ to the loss function (which may be useful when training classes are imbalanced),
the quadratic regularizer $\vct \beta^{\T} \vct \beta$ controls overfitting, and $\lambda$ is a regularization parameter. 
Note that sparsity is not induced here by a sparse regularizer on $\vct \beta$, but rather by setting $m$ to a small value. 
This approach is clearly linear in the space induced by the similarity function, but not necessarily in the input space $\set X$, \ie, the similarity mapping can be used to induce nonlinearity as in kernel methods.
The objective function in Problem~\eqref{eq:objective} can be rewritten in matrix form:
\begin{align}
\label{eq:objective-matrix}
\Omega \left(\vct \beta, b, \vct z\right) =  \left( \vct g^{\T} \mat U \vct g -2
\vct y^{\T} \mat U \vct g + \vct y^{\T} \mat U \vct y \right ) 
+ \lambda \vct \beta^{\T} \vct \beta
\enspace ,
\end{align}
where the column vectors $\vct g=\mat S_{\vct x \vct z} \vct \beta + b \in \mathbb R^{n}$ and $\vct y \in \mathbb R^{n}$ consist of the values of $g$ and $y$ for the training data, $\mat U \in \mathbb R^{n \times n}$ is a diagonal matrix such that ${\rm diag}(\mat U) = (u_{1},\ldots,u_{n})$, and $\mat S_{\vct x \vct z} \in \mathbb R^{n \times m}$ is the similarity matrix computed between $\vct x_{1}, \ldots, \vct x_{n}$ and the prototypes $\vct z$. 

The objective function in Eq.~\eqref{eq:objective-matrix} can be iteratively minimized by modifying $\vct \beta$, $b$ and $\vct z$. First, we randomly initialize the prototypes $\{\vct z^{(0)}_{j}\}_{j=1}^{m}$ with $m$ training samples from $\set D$, and then iteratively repeat the two steps described below.

\myparagraph{(1) $\vct \beta$-step.} The optimal coefficients $\vct \beta$ are computed while keeping the prototypes $\vct z$ fixed.
This amounts to solving a standard ridge regression problem, whose analytical solution is given by deriving Eq.~\eqref{eq:objective-matrix} with respect to $\vct \beta$ and $b$ (with $\vct z$ constant), and then setting the corresponding gradients to zero:
\begin{equation}
\underbrace{\begin{bmatrix}
\mat S^{\T}_{\vct x \vct z} \mat U \mat S_{\vct x \vct z} + \lambda \mathbb I  && \mat S^{\T}_{\vct x \vct z} \mat U \vct 1 \\
\vct 1^{\T} \mat U \mat S_{\vct x \vct z} && \vct 1^{\T} \mat U \vct 1
\end{bmatrix}}_{\mat M}
\begin{bmatrix}
\vct \beta \\
b
\end{bmatrix}
=
\begin{bmatrix}
\mat S^{\T}_{\vct x \vct z} \\
\vct 1^{\T}
\end{bmatrix}
\mat U \vct y \enspace ,
\label{eq:beta}
\end{equation}
where $\mathbb I \in \mathbb R^{m \times m}$ denotes the identity matrix.
Note that the system given by Eq.~\eqref{eq:beta} can be iteratively solved without necessarily inverting $\mat M$, \eg, using stochastic gradient descent~\cite{zhang04-sgd}.

\myparagraph{(2) $\vct z$-step.} If the similarity function is differentiable, Eq.~\eqref{eq:objective-matrix} can be minimized through gradient descent (as no closed-form solution exists for this problem).
Deriving with respect to a given $\vct z_{j}$, we obtain:
\begin{align}
\diff{\Omega}{\vct z_{j}} =  2 \left ( \vct g - \vct y \right )^{\T} \mat U  \left (  \beta_{j}  \diff{\mat S_{\vct x \vct z_{j}}}{\vct z_{j}} + \mat S_{\vct x \vct z} \diff{\vct \beta}{\vct z_{j}}  + \vct 1\diff{b}{\vct z_{j}} \right ) +  
 2 \lambda \vct \beta^{\T} \diff{\vct \beta}{\vct z_{j}} \enspace ,
 \label{eq:gradient-z}
\end{align}
where $\mat S_{\vct x \vct z_{j}}$ is the $j^{\rm th}$ column of $\mat S_{\vct x \vct z}$. Note that all the derivatives computed with respect to $\vct z_{j}$ here are vectors or matrices with the same number of columns as the dimensionality of $\vct z_{j}$.
To compute $\diff{\vct \beta}{\vct z_{j}}$ and $\diff{b}{\vct z_{j}}$, as required by Eq.~\eqref{eq:gradient-z}, we derive Eq.~\eqref{eq:beta} with respect to $\vct z_{j}$ and solve for the required quantities:
\begin{equation}
 \begin{bmatrix}  \diff{\vct \beta}{\vct z_{j}} \\ \diff{b}{\vct z_{j}} \end{bmatrix} = 
- \mat M^{-1} \left ( \beta_{j} \begin{bmatrix} \mat S_{\vct x \vct z}^{\T} \\ \vct 1^{\T} \end{bmatrix} 
+ \begin{bmatrix} \mat V^{\T} \\ \vct 0^{\T} \end{bmatrix} 
\right )
\mat U \diff{\mat S_{\vct x \vct z_{j}}}{\vct z_{j}} \enspace ,
\label{eq:final-gradient}
\end{equation}
where $\mat V \in \mathbb R^{n \times m}$ is a matrix that consists of all zeros except for the $j^{\rm th}$ column, which is equal to $(\vct g- \vct y)$, and $\vct 0, \vct 1 \in \mathbb R^{n}$ are column vectors of respectively $n$ zeros and $n$ ones.

The complete algorithm is given as Algorithm~\ref{alg:support-faces}. 
Note that we invert the $\vct z$- and $\vct \beta$-step detailed above for the sake of compactness. In fact, the coefficients $\vct \beta$ and $b$ should always be updated after changing the prototypes.
Our method can also be used to reduce the number of prototypes used by kernel-based or prototype-based classifiers, like the SVM, by setting the target variables $y$ to the values of the discriminant function of the target classifier for each training point. An example is reported in Fig.~\ref{fig:2D-example}.\footnote{Exploiting the values of a classifier's discriminant function as the target variables $y$ in our approach usually works better than using the true classification labels $y$ directly. The reason is that our approach uses the $\ell_{2}$ loss, which is best suited to regression tasks. For classification, one should indeed exploit a loss function tailored to classification approaches, \eg, the hinge loss.} 
Worth remarking, the virtual prototypes found by our algorithm are quite different from the SVs found by the SVM. In fact, the latter are close to the boundary of the discriminant function, whereas our prototypes are found approximately at the centers of small clusters of training points.

\myparagraph{Gradient of $s(\vct x_{i}, \vct z_{j})$.} In Eq.~\eqref{eq:final-gradient}, the computation of the derivative of $s(\vct x_{1}, \vct z_{j})$, \ldots, $s(\vct x_{n}, \vct z_{j})$ with respect to the corresponding $\vct z_{j}$, depends on the given similarity measure $s$. If $s$ has an analytical representation, like in the case of kernels, the derivative can be easily computed; \eg, for the RBF kernel, $s(\vct x_{i}, \vct z_{j})=\exp(-\gamma ||\vct x_{i} - \vct z_{j}||^{2})$, and  $\diff{s(\vct x_{i}, \vct z_{j})}{\vct z_{j}}= 2\gamma\exp(-\gamma ||\vct x_{i} - \vct z_{j}||^{2}) (\vct x_{i} - \vct z_{j})$.
Otherwise, the gradient can be only approximated numerically, by querying $s(\cdot, \vct z_{j})$ in a neighborhood of $\vct z_{j}$. This is computationally costly, especially if $\vct z_{j}$ is high dimensional.
In the case of images, we have found that the similarity tends to increase while shifting $\vct z_{j}$ towards $\vct x_{i}$, even if this shift is operated linearly in the space of the pixel values (\eg, by computing a convex combination of the two images)~\cite{biggio15-icb,demontis15-iciap,melis15-iciap}. This amounts to approximating the gradient as
\begin{align}
\diff{s(\vct x_{i}, \vct z_{j})}{\vct z_{j}} = s(\vct x_{i}, \vct z_{j}) \left ( \vct x_{i} - \vct z_{j} \right ) \enspace .
\label{eq:similarity-grad}
\end{align}
Although using this heuristic to approximate the gradient of $s$ may affect, in general, the convergence of our algorithm, in the next section we show that it works quite well even when the similarity function is not analytically given (\ie, for a graph-based face matching algorithm). Clearly, different heuristics may be considered, if the proposed one turns out to be not suited to the task at hand.

\begin{algorithm}[t] 
  \caption{Super-sparse Learning Machine}
  \label{alg:support-faces}
  \textbf{Input:} The training data $\set D =\{\vct x_{i}, y_{i}\}_{i=1}^{n}$; the similarity function $s(\cdot,\cdot)$; the regularization parameter $\lambda$; the initial coefficients $(\vct \beta^{(0)}, b^{(0)})$ and vectors $\{\vct z^{(0)}_{j}\}_{j=1}^{m}$; the gradient step size $\eta$; a small number $\varepsilon$.\\
  \textbf{Output:} The learned coefficients $(\vct \beta, b)$ and vectors $\{\vct z_{j}\}_{j=1}^{m}$.
{\fontsize{11}{10}\selectfont
  \begin{algorithmic}[1]
    \STATE{Set the iteration count $t \gets 0$.}
    \REPEAT
    \STATE{Set $j \gets {\rm mod}(t,m)+1$ to index a virtual prototype.}
    \STATE{Increase the iteration count $t \gets t + 1$}
      \STATE{\textbf{$\vct z$-step.} Set $\vct z_{j}^{(t)} \gets  \vct z_{j}^{(t-1)} - \eta \diff{\Omega}{\vct z_{j}^{(t-1)}} - \frac{1}{t^{2}}\diff {\mat S_{\vct z \vct z_{j}}}{\vct z_{j}^{(t-1)}}$ (Eq.~\ref{eq:gradient-z} with  penalty for close prototypes).}
      \STATE{Project $\vct z_{j}^{(t)}$ onto the feasible domain, if any (\eg, a box constraint).}
      \STATE{Set $\vct z_{i}^{(t)} = \vct z_{i}^{(t-1)}$, $\forall i \neq j$.}
      \STATE{\textbf{$\vct \beta$-step.} Compute $(\vct \beta^{(t)}, b^{(t)})$ using $\vct z^{(t)}_{1}, \ldots, \vct z^{(t)}_{m}$ (Eq.~\ref{eq:beta}).}
      \UNTIL{$\left | \Omega (  \vct \beta^{(t)},  b^{(t)}, \vct z^{(t)} ) - \Omega ( \vct \beta^{(t-1)},  b^{(t-1)}, \vct z^{(t-1)} ) \right | < \varepsilon$}
      \STATE{\textbf{return:} $\vct \beta = \vct \beta^{(t)}$, $b = b^{(t)}$ and $\vct z = \vct z^{(t)}$.}
  \end{algorithmic}
 }
\end{algorithm}

\myparagraph{Prototype Initialization.} 
Our approach might suffer from the intrinsic nature of the non-convex optimization problem faced in the $\vct z$-step, \ie, when optimizing the virtual prototypes. In fact, due to the presence of multiple local minima (in which some prototypes may be too close to each other), our algorithm turns out to be quite sensible to the initialization of the virtual prototypes~\cite{biggio15-icb,demontis15-iciap,melis15-iciap}.
To overcome this limitation, we propose the following strategy. Instead of running the algorithm multiple times with different initializations (which would increase the overall computational complexity of our approach), we modify the gradient defined in Eq.~\eqref{eq:gradient-z} to account for a penalty term. It aims to reduce similarities between virtual prototypes, avoiding them to converge towards the same points. The penalty that we add to the gradient of Eq.~\eqref{eq:gradient-z} is simply $\diff {\mat S_{\vct z \vct z_{j}}}{\vct z_{j}}$.
Since this term achieves large values for prototypes that are closer to $\vct z_{j}$, it is clear that $\vct z_{j}$ will be shifted away from them during optimization.
We further multiply the penalty term by a decaying coefficient (\eg, $t^{-2}$, being $t$ the iteration count) to distance the prototypes sufficiently during the first iterations of the algorithm, without affecting convergence of our algorithm.

\myparagraph{Selecting the number of virtual prototypes.} Another important issue in our method regards the selection of $m$. As already discussed, $m$ can be directly defined from specific constraints; otherwise, when some some degree of flexibility is allowed, it is possible to tune $m$ by minimizing the number of prototypes without significantly compromising accuracy.
To this end, we define an objective function characterizing this trade-off as
\begin{equation}
\label{eq:xval-m}
\set L(m)= \ell(m) + \rho \cdot m \, ,
\end{equation} 
and set $m = \textstyle \argmin_{m^{\prime}} \; \set L(m^{\prime})$. In the above objective, $\ell$ is a loss (error) measure to be evaluated on a validation set, and $\rho$ is a trade-off parameter. For higher $\rho$ values, fewer prototypes are selected, at the expense of higher error rates.

\begin{figure*}[t]
\linespread{1}
\centering
\includegraphics[width=\textwidth]{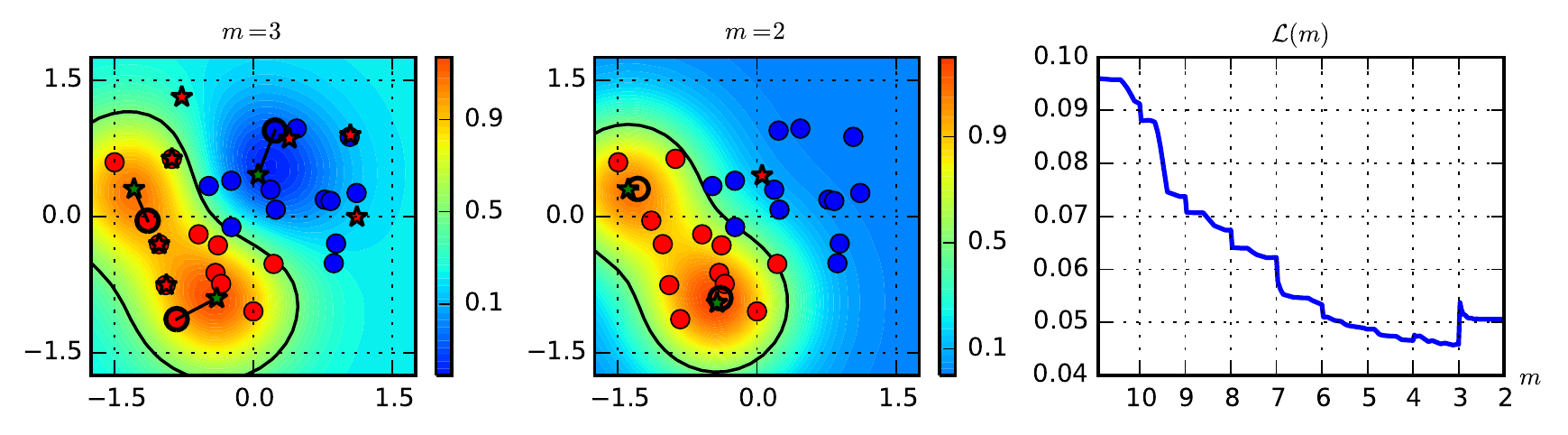}
\caption{An example of our incremental CV procedure to select $m$.
We initially set $m=10$, and iteratively remove one prototype at a time, up to $m=2$.
The discriminant function $g(\vct x)$  of our super-sparse learning algorithm is shown (in colors) in the left and middle plot, for $m=3$ and $m=2$. The initial and final positions of the current prototypes are shown as black circles and green stars, respectively, while the prototypes removed at the previous iterations are reported as red stars. The minimum of $\set L(m)$ (Eq.~\ref{eq:xval-m}) is attained for $m=3$ (using  the mean squared error as the loss function $\ell(m)$, and $\rho=10^{-3}$), as shown in the left plot. In fact, for $m=2$, the set of blue points is not properly represented, and $\set L(m)$ increases.}
\label{fig:xval-m}
\end{figure*}

The value of $m$ that minimizes Eq.~\eqref{eq:xval-m} can be efficiently selected through an incremental CV procedure which uses a grid search on some pre-defined values $m \in \{m_{j}\}_{j=1}^{K}$, as described in what follows. Let us assume that $m_{1} > \ldots > m_{K}$. 
We first learn our approach using the largest number of prototypes $m_{1}$. Then, to learn the solution with $m_{2}$ prototypes, we remove the $m_{1}-m_{2}$ prototypes assigned to the smallest $\boldsymbol{\beta}$ coefficients (in absolute value), and update the remaining $m_{2}$ coefficients and prototypes by re-running the learning algorithm from the current solution (warm start).
We iterate until the most compact solution with $m_{K}$ prototypes is learned, and then select the value of $m$ that minimizes Eq.~\eqref{eq:xval-m}. An example is given in Fig.~\ref{fig:xval-m}.

\myparagraph{Computational Complexity.} As previously discussed, our learning algorithm consists of two steps. The $\vct \beta$-step is computationally lightweight, as it only requires solving a linear system involving $m +1$ variables. Furthermore, this system has to be solved from scratch only at the first iteration, while for subsequent iterations one can exploit the previous solution as a warm start, and use an iterative algorithm to converge to the solution very quickly.
The most computationally-demanding part is instead the computation of $\mat S_{\vct x \vct z_{j}}$ during the $\vct z$-step, which has complexity of $\set O(n)$, as we optimize one prototype at a time.
Clearly this has to be repeated at each iteration. However, as our approach typically converges within 20 to 30 iterations, it is likely that it remains faster than other prototype-based learning algorithms. The reason is that the latter usually require computing the entire similarity matrix, which costs $\set O(n^{2})$ operations. We will nevertheless discuss some techniques to reduce the training complexity of our approach in Sect.~\ref{sect:contributions_and_future_work}.
As for the complexity at test time, it is clear that our approach drastically reduces it to $\set O(m)$ operations.

\myparagraph{Structured Inputs.} Worth remarking, if the prototypes are not represented in terms of feature vectors, but using more complex structures (\eg, graphs or strings), then the $\vct z$-step of our algorithm can not be optimized through gradient descent. More generally, one may define a set of minimal modifications to each prototype (\eg, adding or removing a vertex in a graph, or characters in a string), and select those that greedily minimize the objective in Eq.~\eqref{eq:objective-matrix}. Note however that this black-box optimization procedure is clearly computationally demanding, and further empirical investigations are required to validate its effectiveness and extend our approach to more generic, structured inputs.

\section{Application Examples} 
We report here two application examples related to face verification and age estimation from faces~\cite{biggio15-icb,demontis15-iciap}.
The goal of these examples is to show how and to what extent our super-sparse reduction approach can improve computational efficiency at test time without almost affecting system performance.

\subsection{Biometric Identity Verification from Faces} 

Face verification consists of validating if a client is claiming his/her real identity (genuine claim) or he/she is pretending to be someone else (impostor claim). To this end, his/her face is acquired through a camera and compared against the reference prototypes of the claimed identity.
To save time and memory, few reference prototypes are used.
The corresponding similarity values are then combined with heuristic schemes, separating prototype selection from the algorithm used to combine the similarities~\cite{biggio15-icb}.

We train a one-vs-all SVM for each client.
This automatically selects the optimal prototype gallery (\ie, the SVs), which however is often too large.
We show that the proposed algorithm can successfully reduce the number of SVs, without affecting the recognition performance.
We use the benchmark AT\&T\footnote{\url{http://www.cl.cam.ac.uk/research/dtg/attarchive/facedatabase.html}} and BioID\footnote{\url{https://www.bioid.com/About/BioID-Face-Database}} face datasets, respectively consisting of 40 clients with 10 face images each, and of 1,521 face images belonging to 23 clients.
We assume that half of the clients are enrolled into the system, while the remaining ones are used only as impostors at test time.
For training we randomly select 5 face images per enrolled client, and simulate impostors using face images of enrolled clients that do not belong to the claimed identity.
At test time, we simulate impostors using the non-enrolled clients, and genuine claims using all the remaining face images of the claimed identity not used in the training set. This guarantees that the impostors are different between training and test sets (a common practice when testing biometric systems).
Results are averaged over five different training-test pairs and client splits.

\myparagraph{Matching algorithms.} We use two different matching algorithms to compute similarities between faces.

\noindent \emph{1) Eigenface-based RBF Kernel}~\cite{turk91,jonsson02}. This algorithm maps each face image onto a reduced $d$-dimensional feature vector using principal component analysis (PCA). 
We select $d$ so as to preserve 95\% of the variance of the data. We then use the RBF kernel $s(\vct x_{i}, \vct x_{j}) = \exp ( {-\gamma \| \vct x_{i} -  \vct x_{j} \|}^2 )$ as the similarity measure, and set $\gamma = 1/d$~\cite{libsvm}. On average, we got $d=56$ for AT\&T, and $d=41$ for BioID.

\noindent \emph{2) Elastic Bunch Graph Matching (EBGM)}~\cite{beveridge05}. 
It extracts a bunch of image descriptors (\ie, Gabor wavelets) at predefined landmark points in face images (\eg, eyes and nose). These points correspond to nodes in a graph, which is elastically deformed to account for different face orientations and expressions during matching. In this case, $s(\vct x_{i}, \vct x_{j})$ is neither analytically given, nor positive semi-definite.

\begin{figure*}[t]
\linespread{1}
\centering
\includegraphics[width=0.425\textwidth]{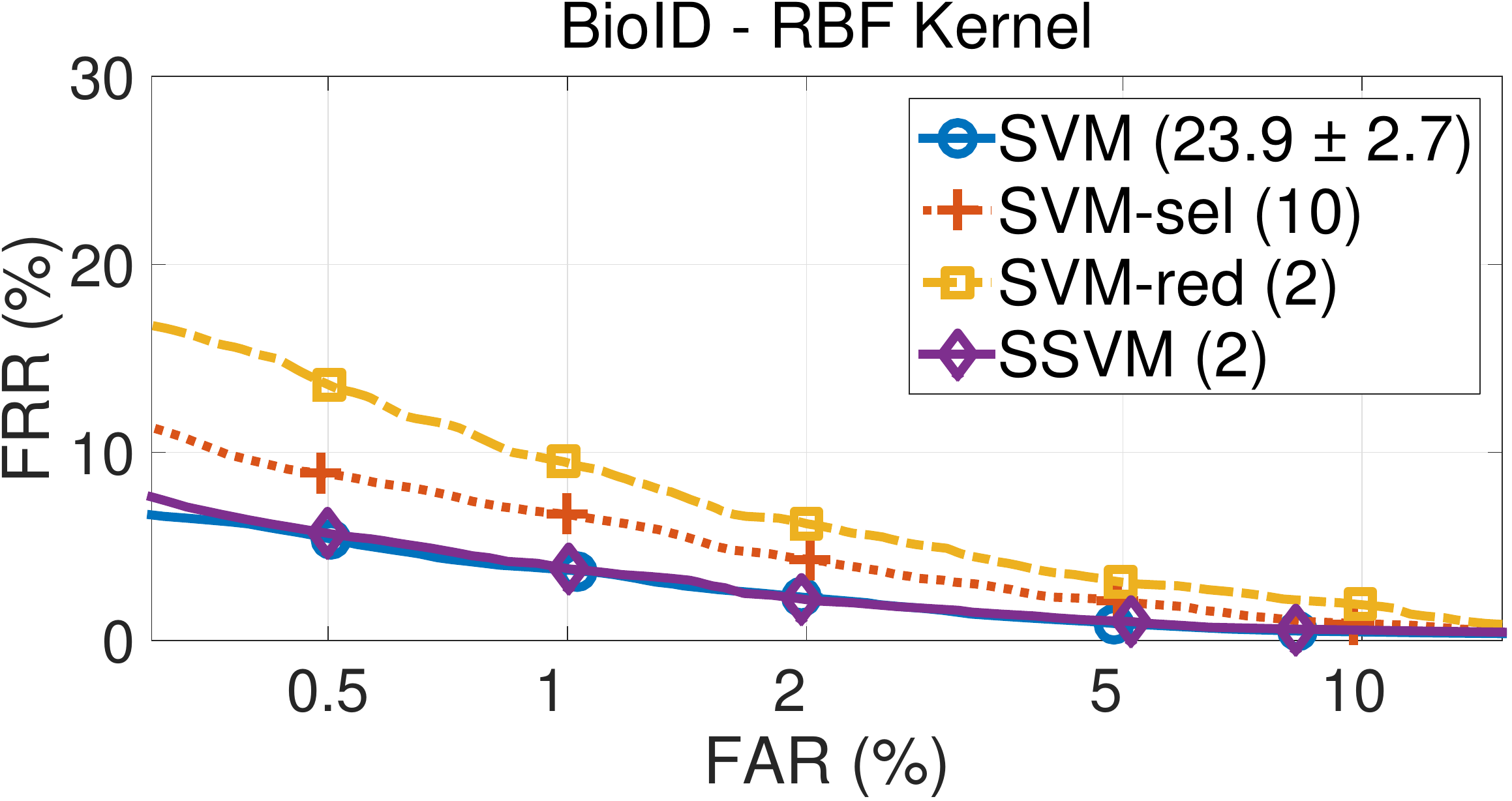} \enspace
\includegraphics[width=0.425\textwidth]{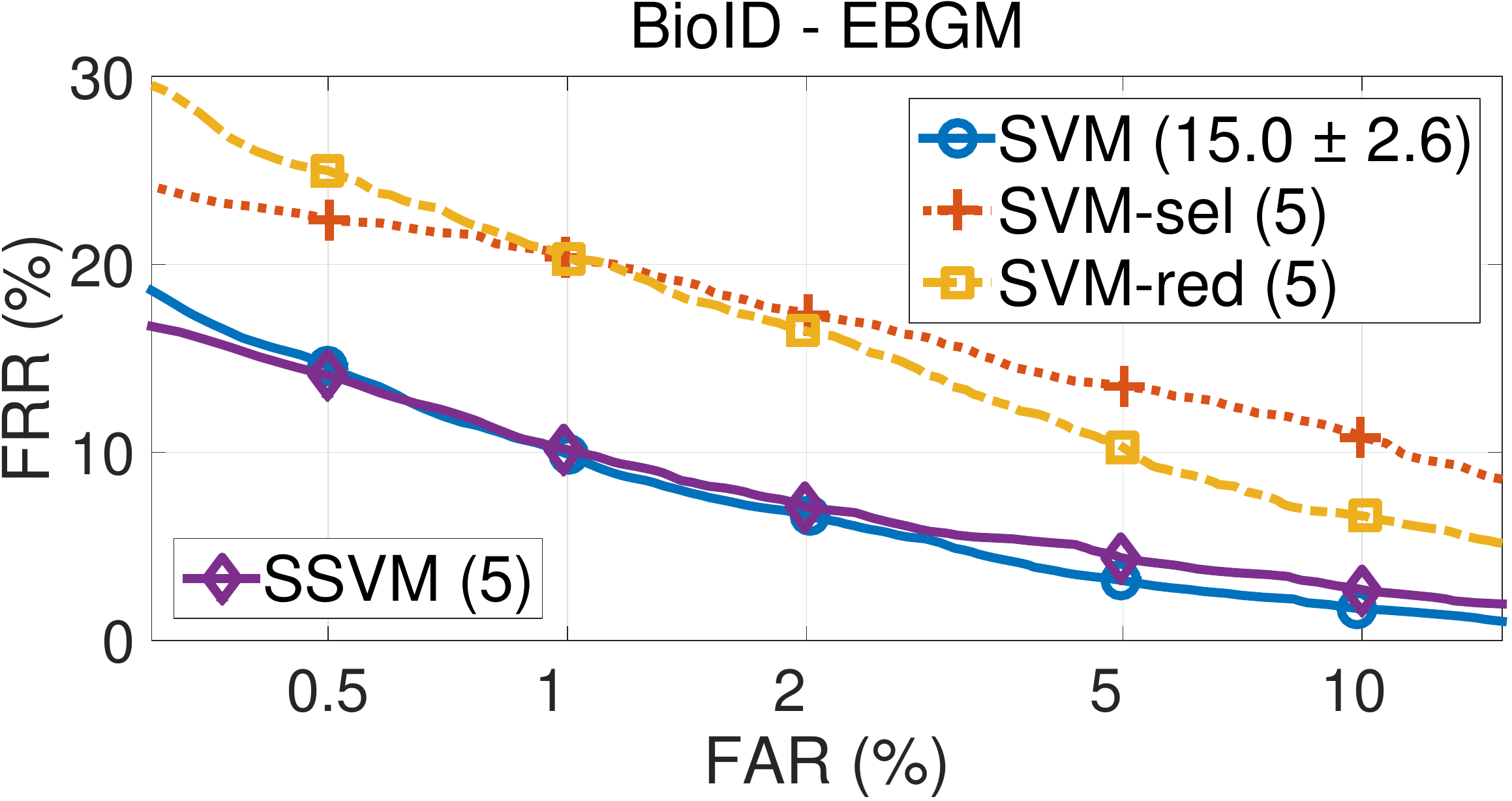}\\\vspace{0.5em} \includegraphics[width=0.425\textwidth]{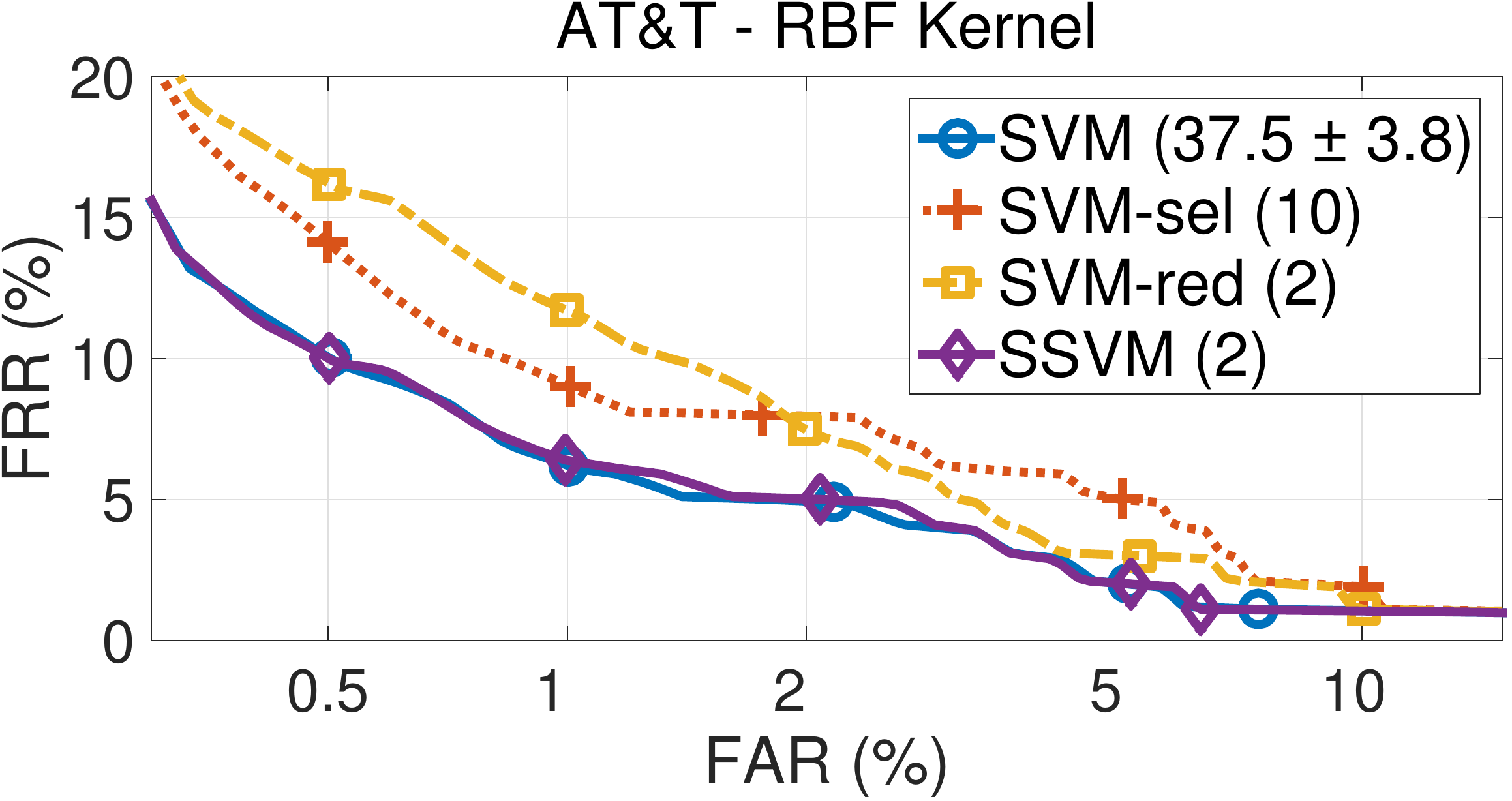} \enspace
\includegraphics[width=0.425\textwidth]{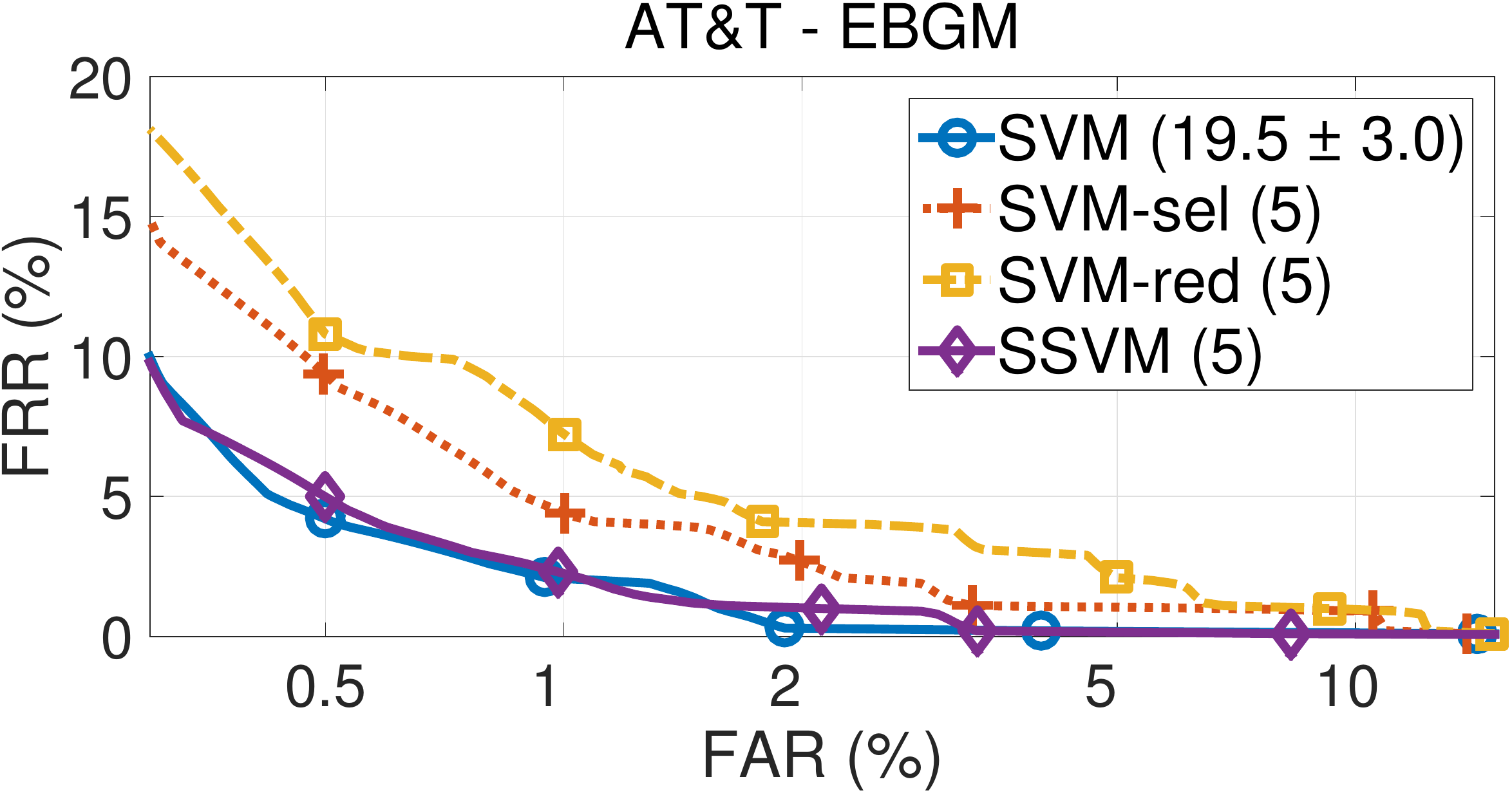} 
\caption{Averaged FRR vs FAR values for the BioID (\emph{top row}) and AT\&T (\emph{bottom row}) face datasets, and for the Eigenface-based RBF Kernel (\emph{left column}) and EBGM (\emph{right column}) algorithms. The average number of matchings required for verification and the std. dev. is reported in parentheses.}
\label{fig:perf-ROC}
\end{figure*}

\myparagraph{Verification methods.} We compare our approach to learn super-sparse SVMs (\ssfm) with the \svmsel{} and \svmred{} techniques for reducing the number of SVs~\cite{scholkopf99} (see Sect.~\ref{sect:related-work}), and also consider the standard \svm{} for comparison.

We set the SVM regularization parameter $C \in \{10^{-1}, 10^{0}, \ldots, 10^{3}\}$ by maximizing recognition accuracy through a 5-fold CV. As the two classes are highly unbalanced, we used a different $C$ value for each class, multiplying it by the prior probability of the opposite one, estimated from training data.
To ensure fast convergence of our Algorithm~\ref{alg:support-faces} to a stable solution, we run a set of preliminary experiments and set the gradient step $\eta=0.5$ for both datasets; the regularization factor $\lambda=10^{-6}$ for the Eigenface-based RBF Kernel, and $\lambda=10^{-3}$ and $\lambda = 10^{-5}$ for the EBGM on the BioID and AT\&T data, respectively.
The gradients of $s(\vct x_{i}, \vct z_{j})$ are analytically computable for the RBF Kernel; for the EBGM (which is not given analytically) we use the approximate gradient given in Eq.~\eqref{eq:similarity-grad}.

\myparagraph{Results.} Fig.~\ref{fig:perf-ROC} shows the fraction of incorrectly-rejected genuine claims (false rejection rate, FRR) vs. the fraction of incorrectly-accepted impostors (false acceptance rate, FAR) for each method, obtained by varying each client-specific threshold, and then by averaging over all clients and repetitions. The average number of matchings (prototypes) required by each method at test time is also reported.
Except for \svm{}, this number has to be fixed in advance: for \svmsel{}, \svmred{}, and \ssfm{} we respectively set it to 10, 2 and 2, when using the RBF Kernel, and to 5, for all methods, when using the EBGM.

Our \ssfm{} achieves comparable performance as \svm{} but using only 2 and 5 virtual prototypes (instead of more than 20 and 15), respectively, for the RBF kernel and EBGM matching algorithms. Conversely, \svmsel{} and \svmred{} require a higher number of prototypes to achieve a comparable performance. This highlights that a principled approach may guarantee high recognition accuracy and efficiency using an extremely sparse set of virtual prototypes, without even knowing analytically the matching algorithm.

\begin{figure*}[t]
\linespread{1}
\centering
\minipage{0.475\textwidth}
\centering
\includegraphics[width=0.975\textwidth]{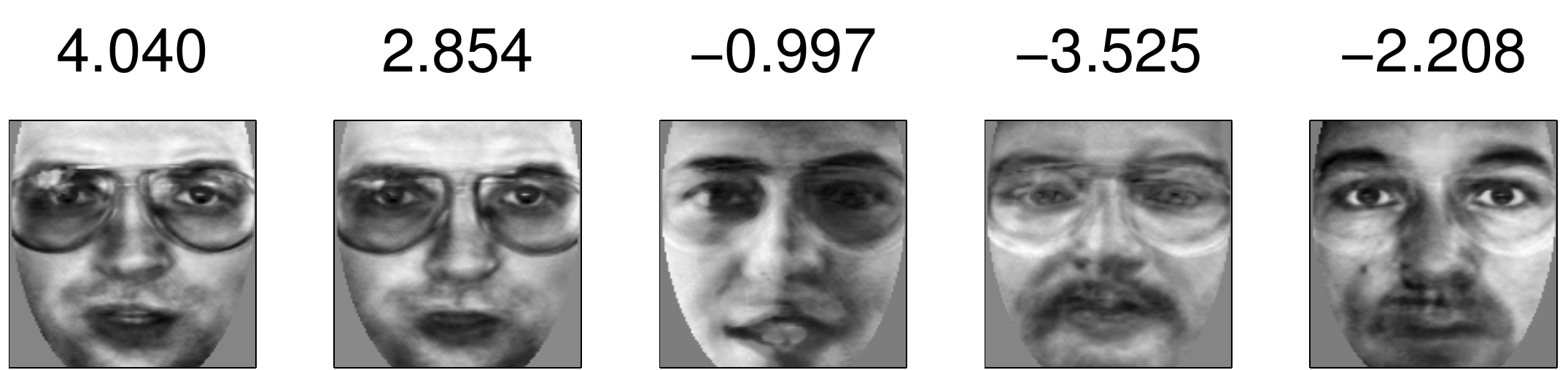}
\caption{Prototypes and $\vct \beta$ values learned by our \ssfm{} for a client in the BioID dataset, using the EBGM matching algorithm.}
\label{fig:bioID-faces19-EBGM}
\endminipage\quad
\raisebox{0.8em}{
\minipage{0.475\textwidth}
\centering
\includegraphics[width=0.18\textwidth]{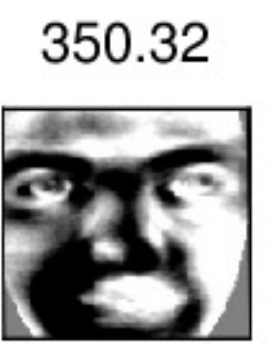}
\includegraphics[width=0.18\textwidth]{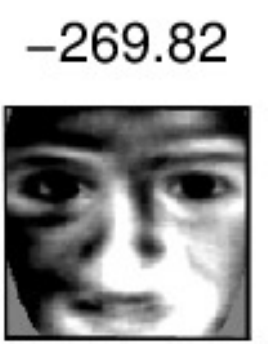}
\includegraphics[width=0.18\textwidth]{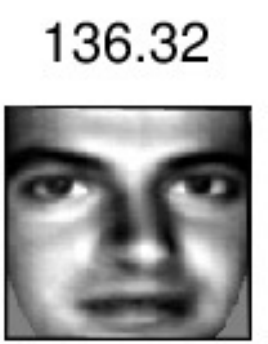}
\includegraphics[width=0.18\textwidth]{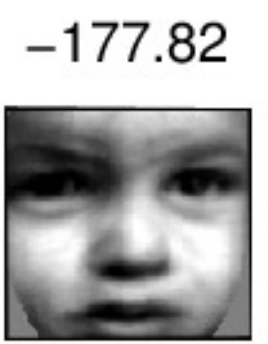}
\caption{Prototypes and $\vct \beta$ values learned by our super-sparse LASSO from the Fg-Net data.}
\label{fig:interpretability}
\endminipage}
\end{figure*}

\myparagraph{Interpretability.}
An example of the virtual prototypes found by our \ssfm{} is shown in Fig.~\ref{fig:bioID-faces19-EBGM}.
We can appreciate that the genuine (virtual) prototypes ($\vct \beta > 0$) are obtained by merging genuine prototypes, preserving the aspect of the given client.
Impostor prototypes ($\vct \beta < 0$) are instead the combination of faces of different identities, to compactly represent information about impostors. 
Although these prototypes do not correspond to any real user, interestingly they still resemble face images. 
This makes our \ssfm{} approach interpretable, in the sense that a face image is considered genuine if it is sufficiently similar to the genuine prototypes found by our algorithm (and different from the impostor ones).

\subsection{Age Estimation from Faces} 
The goal here is to predict the age of a person from a photograph of his/her face. 
We tackle this problem as a regression task, as most of the existing methods, and show that our approach can be helpful in this context to: ($i$) speed up age estimation at test time by dramatically reducing the number of reference prototypes; ($ii$) provide more interpretable decisions; and ($iii$) mitigate the risk of overfitting to specific face databases. The experimental setup is similar to the one defined in our previous work~\cite{demontis15-iciap}.

\begin{figure*}[t]
\linespread{1}
\centering
\includegraphics{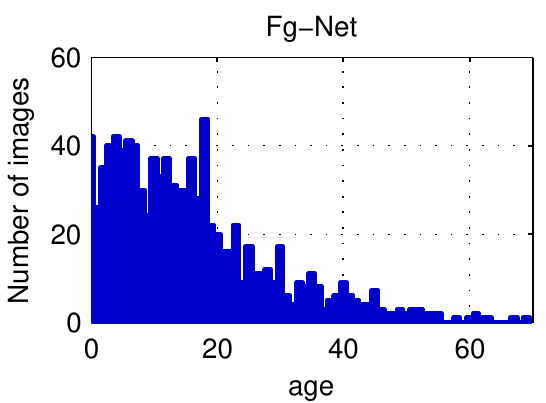}
\hspace{0.25em}
\includegraphics{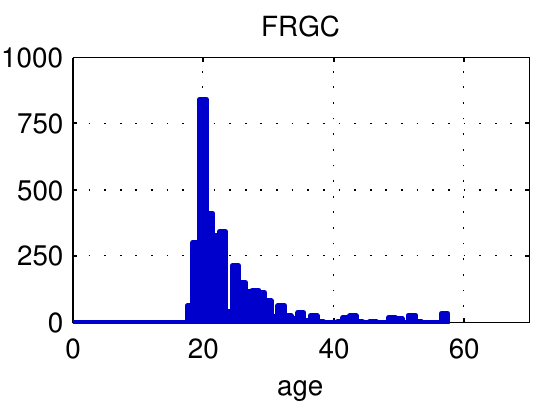}
\caption{Number of images per age (years) for Fg-Net (\emph{left}) and FRGC (\emph{right}).}
\label{fig:dataset-ages}
\end{figure*}

\myparagraph{Datasets.} We use two publicly-available benchmark face databases: Fg-Net Aging and FRGC.
Fg-Net is the main database for this task.
It includes about 1,000 images of 82 subjects acquired in a totally uncontrolled condition, which makes it particularly challenging.
Many images are blurred, exhibit different resolutions and illumination conditions, and the number of subjects per age is not equally distributed. The age range for each subject varies from 0 to 69 years, although the majority of images belong to 20-year-old people. 
FRGC consists of about 50,000 face images acquired in different time sessions, belonging to about 500 people (about 200 females and 300 males) of different ethnicity, with ages spanning from 17 to 69 years.
Face images were acquired in a controlled indoor environment, in frontal pose, to facilitate the recognition task. To keep the complexity of our experiments manageable, we restrict our analysis to a subset of about 5,000 images, randomly selected from this dataset. The age distributions of both datasets are shown in Fig.~\ref{fig:dataset-ages}.

\myparagraph{Experimental Setup.} 
We normalize images as discussed in~\cite{demontis15-iciap}, and reduce the resulting set of 19,500 features (\ie, pixel values) through linear discriminant analysis (LDA), retaining the maximum number of components (\ie, the number of different age values minus one).
We evaluate performance in terms of Mean Absolute Error:
${\rm MAE}=\frac{1}{r}\sum_{i=1}^{r}  \left | g(\vct x_{i}) - y_{i} \right |$,
where $g(\vct x_{i})$ is the regression estimate of our approach for the $i^{\rm th}$ subject, whose true age is $y_{i}$, and $r$ is the number of test images.
We average results using a 5-fold CV procedure where each subject appears only in one fold.  
We use the RBF kernel as the similarity measure. 
We consider LASSO~\cite{tibshirani96-lasso} and ridge~\cite{hoerl70-ridge} regression, and optimize their regularization parameter through CV.
We compare our approach against the following prototype-selection methods~\cite{riesen07,spillmann06-spr}:
Random (PS-R), which randomly selects $m$ prototypes from the training data;
Border (PS-B), which selects $m$ prototypes from the frontier of the training data;
Spanning (PS-S), which selects the first prototype as the training-set median, and the remaining ones through an iterative procedure that maximizes the distance to the set of previously-selected prototypes;
and $k$-medians (PS-KM), which runs $k$-means clustering to obtain $m$ clusters from the training set, and then selects the $m$ prototypes as their set medians (\ie, the medians of the clusters).
We optimize $m$ according to the CV procedure defined in Sect.~\ref{sect-super-sparse-virtual-vector-machine}, using the MAE as the loss function $\ell(m)$ and $\rho=0.1$ (Eq.~\ref{eq:xval-m}).
We consider also a cross-database scenario in which training and test sets are drawn from different databases, to verify if prototype-selection methods are less prone to overfitting.
	
\begin{table*}[t] 
\linespread{1}
\caption{Average MAE and the number of prototypes (in parentheses) selected by Ridge, LASSO, and the corresponding prototype-selection methods, for LDA-based features. Each column reports training/test sets, including cross-database evaluations; \eg, Fg-Net/FRGC means that Fg-Net is used for training, and FRGC for testing. The lowest error values are highlighted in bold, for each configuration.}
\fontsize{8}{12}\selectfont
	\centering 
	{
		\begin{tabular}{| p{2.2 cm} | p{2cm} | p{2cm} | p{2cm} | p{2cm} | }  
			\hline  
			\textbf{Method} & \textbf{Fg-Net/Fg-Net} & \textbf{Fg-Net/FRGC} & \textbf{FRGC/Fg-Net} & \textbf{FRGC/FRGC} \\ \hline 
			\hline  
Ridge  &8.00 (781.6)  &8.46 (781.6)  &14.85 (2747.2)  &4.53 (2747.2) \\  
PS-R Ridge &9.93 (5.0)  &9.11 (5.0)  &\textbf{12.98} (4.0)  &\textbf{4.10} (4.0) \\  
PS-B Ridge &36.78 (5.0)  &30.48 (5.0)  &26.34 (4.0)  &17.35 (3.2) \\  
PS-S Ridge  &11.13 (5.0)  &10.36 (5.0)  &15.29 (4.0)  &4.85 (4.0) \\  
PS-KM Ridge &10.45 (5.0)  &13.94 (5.0)  &13.62 (4.0)  &4.04 (4.0) \\  
SRidge &\textbf{9.06} (5.2)  &\textbf{7.96} (4.6)  &14.42 (4.4)  &4.31 (4.2) \\  
\hline   
LASSO &7.92 (60.0)  &8.99 (60.2)  &14.71 (20.8)  &4.67 (22.4) \\  
PS-R LASSO &11.77 (5.0)  &\textbf{7.81} (7.0)  &14.54 (3.0)  &5.40 (2.4) \\  
PS-B LASSO &36.78 (5.2)  &28.66 (7.0)  &25.80 (2.6)  &17.48 (2.2) \\  
PS-S LASSO &10.54 (5.0)  &10.34 (7.2)  &\textbf{14.15} (3.0)  &\textbf{4.48} (3.0) \\  
PS-KM LASSO &12.54 (5.0)  &9.96 (7.0)  &15.37 (3.0)  &5.14 (3.0) \\  
SLASSO &\textbf{7.99} (6.2)  &9.09 (7.6)  &14.76 (3.6)  &4.75 (4.0) \\  
\hline   
\end{tabular}}\label{tab:mae}
\end{table*}

\myparagraph{Results.} As shown in Table~\ref{tab:mae}, both for the standard and cross-database evaluations, our algorithms exhibit almost the same performance as their non-sparse versions, despite using fewer prototypes. They often outperform also the competing prototype-selection methods, or achieve a comparable performance.

\myparagraph{Interpretability.} Interpretability of decisions is important to determine whether meaningful aging patterns are learned, \ie, if the age of a subject can be correctly predicted from face images of different datasets. Fig.~\ref{fig:interpretability} shows a set of prototypes found by our super-sparse LASSO algorithm, which correctly assigns higher $\vct \beta$ values to older people.

\section{Related Work} 
\label{sect:related-work}

Besides work in similarity-based learning, discussed in Sects.~\ref{sect:introduction}-\ref{sect:learning-similarity-spaces}, another line of research related to super-sparse learning in similarity spaces is the one related to SVM reduction approaches, \ie, approaches that aim at reducing the number of SVs of an SVM, or to learn SVM with reduced complexity directly~\cite{scholkopf99,keerthi06,wang12b}.
Worth remarking, however, our approach is not specifically designed for SVMs and kernel machines, as it can be in principle applied to generic similarity functions and learning algorithms.
Moreover, the first reported application example on face verification has also demonstrated that some of the existing methods for reduction of SVs in SVMs (\ie, \svmsel{} and \svmred{}) can not achieve reduction rates and accuracy that are comparable with those achieved by our method. These methods in practice reduce the number of SVs by minimizing the $\ell_{2}$ distance computed between the hyperplane normal of the given (unpruned) SVM and that of the reduced SVM in kernel space, as a function of the dual coefficients $\vct \alpha$ and the set of reduced prototypes~\cite{scholkopf99}.
While $\vct \alpha$ can be analytically found, as in our method, the choice of the reduced prototypes is different: \svmsel{} eliminates one prototype at each iteration from the initial set using Kernel PCA to drive the selection; and \svmred{} creates a new prototype at each iteration by minimizing the aforementioned $\ell_{2}$-norm distance.
Both approaches are thus greedy, as the reduced set of prototypes is constructed iteratively by removing or adding a prototype at a time, up to the desired number $m$.
The reason of the superior performance exhibited by our algorithm is thus twofold: (i) \svmsel{} and \svmred{} require the matching algorithm to be a positive semi-definite kernel to \emph{uniquely} find the coefficients $\vct \alpha$;\footnote{In fact, the notion of hyperplane exploited in their objective is only consistent for positive semi-definite kernels.}  (ii) they do not modify the prototypes that are already part of the reduced expansion, and do not even reconsider the discarded ones. Our approach overcomes such limitations by optimizing a different objective (suited to non-positive semi-definite kernels too) and by iteratively modifying the virtual prototypes during the optimization. These may be common advantages also with respect to more recent reduction methods, but this deserves further investigation~\cite{keerthi06,wang12b}.

\section{Summary and Open Problems} 
\label{sect:contributions_and_future_work}

The proposed approach aims to tackle computational complexity of similarity-based classifiers at test time.  Our approach builds on~\cite{biggio15-icb,demontis15-iciap,melis15-iciap}, where we originally define our super-sparse learning machines.
Here we have further extended our approach especially from an algorithmic viewpoint, by including a penalty term to reduce sensitivity to initialization of the virtual prototypes, and by designing a specific CV procedure to tune the number of prototypes $m$.
We remark that we do not consider multi-class classification problems, but that an extension of our approach to deal with them has already been developed, exhibiting outstanding performance on image classification~\cite{melis15-iciap}.

Our future research directions aim at designing very efficient (and interpretable) machines at test time.
We will first explore different possibilities to overcome the computational bottleneck of our approach during training; \eg, similarities between the virtual prototypes and the training samples can be only computed after a given number of iterations $p > 1$, instead of being computed at each shift of a virtual prototype. During the intermediate steps, the similarity values can be updated using a first-order approximation, which is provided by the computations performed at the previous iterations.
Another interesting research direction consists of exploiting our super-sparse reduction approach to reduce complexity of other non-parametric estimators, like kernel density estimators and $k$-nearest neighbors.
Finally, it would be interesting also to extend our approach to handle complex input structures like graphs and strings, considering efficient black-box optimization techniques.


\end{document}